\documentclass[letterpaper]{article}
\usepackage{uai2018}
\usepackage[margin=1in]{geometry}

\usepackage{times}

\newcommand{\figspace}{\vspace{-2mm}}

\usepackage{amsmath}
\usepackage{amsfonts}
\usepackage{booktabs}
\usepackage{tikz}
\usetikzlibrary{bayesnet}
\usepackage{graphicx}
\usepackage[caption=false]{subfig}
\usepackage[]{algorithm2e}
\usepackage{algpseudocode}
\usepackage{wrapfig}

\newcommand{\modelname}{\textnormal{ML-GP}}
\newcommand{\baseline}{\textnormal{SGP}}

\newcommand{\x}{\boldsymbol{x}}
\newcommand{\y}{\boldsymbol{y}}
\newcommand{\zi}{\boldsymbol{z}}
\newcommand{\Zi}{\boldsymbol{Z}}
\newcommand{\tx}{\tilde{\boldsymbol{x}}}
\newcommand{\cs}{\boldsymbol{c}}
\newcommand{\h}{\boldsymbol{h}}
\newcommand{\hall}{\boldsymbol{H}}
\newcommand{\f}{\boldsymbol{f}}
\newcommand{\ui}{\boldsymbol{u}}
\newcommand{\Ui}{\boldsymbol{U}}

\newcommand{\m}{\boldsymbol{m}}
\newcommand{\n}{\boldsymbol{n}}
\newcommand{\Sc}{\boldsymbol{S}}
\newcommand{\Tc}{\boldsymbol{T}}

\newcommand{\I}{\boldsymbol{I}}

\newcommand{\KL}{\textnormal{KL}}
\newcommand{\E}{\mathbb{E}}

\newcommand{\mat}[1]{\boldsymbol{#1}}
\renewcommand{\vec}[1]{\boldsymbol{#1}}

\newcommand{\best}[1]{\textbf{\color{blue}{#1}}}
\newcommand{\blue}[1]{\color{blue}{#1}}

\title{Meta Reinforcement Learning 
with\\ Latent Variable Gaussian Processes}

\author{ {\bf Steind\'or S\ae mundsson} \\
Department of Computing\\
Imperial College London\\
United Kingdom
\And
{\bf Katja Hofmann}  \\
Microsoft Research\\
Cambridge\\
United Kingdom
\And
{\bf Marc Peter Deisenroth}  \\
Department of Computing\\
Imperial College London\\
United Kingdom
}

\begin{document}

\maketitle

\begin{abstract}
Learning from small data sets is critical in many practical applications where data collection is time consuming or expensive, e.g., robotics, animal experiments or drug design. Meta learning is one way to increase the data efficiency of learning algorithms by generalizing learned concepts from a set of training tasks to unseen, but related, tasks. Often, this relationship between tasks is hard coded or relies in some other way on human expertise. In this paper, we frame meta learning as a hierarchical latent variable model and infer the relationship between tasks automatically from data. We apply our framework in a model-based reinforcement learning setting and show that our meta-learning model effectively generalizes to novel tasks by identifying how new tasks relate to prior ones from minimal data. This results in up to a $60\%$ reduction in the average interaction time needed to solve tasks compared to strong baselines.
\end{abstract}

\section{INTRODUCTION}
Reinforcement learning (RL) is a principled mathematical framework for learning optimal controllers from trial and error~\cite{Sutton1998}. However, RL traditionally suffers from data inefficiency, i.e., many trials are needed to learn to solve a specific task. This can be a problem when learners operate in real-world environments where experiments can be time consuming (e.g., where experiments cannot run faster than real time) or expensive. For example, in a robot learning setting, it is impractical to conduct hundreds of thousands of experiments with a single robot because we will have to wait for a long time and the wear and tear on the hardware can cause damage.

There are various ways to address data-efficiency in RL. Model-based RL, where predictive models of the transition function are learned from data, can be used to reduce the number of experiments in the real world. The learned model serves as an emulator of the real world. A challenge with these learned models is the problem of model errors: If we learn a policy based on an incorrect model, the policy is unlikely to succeed on the real task. To mitigate the issue of these model errors it is recommended to use probabilistic models and to take model uncertainty explicitly into account during planning~\cite{Schneider1997, Deisenroth2011c}. This approach has been applied successfully to simulated and real-world RL problems~\cite{Deisenroth2015}, where a policy-search approach was used to learn optimal policy parameters. Robustness to model errors and, thereby, increased data efficiency, can be achieved by using model predictive control (MPC) instead of policy search since MPC allows for online updates of the model, whereas policy search would update the model only after a trial~\cite{Kamthe2018}. 

If we are interested in solving a set of related tasks we can use meta learning  as an orthogonal approach to increase data efficiency. Generally, the aim of meta learning is to train a model on a set of training tasks and then generalize to new tasks using minimal additional data~\cite{finn2017model}. The strength of meta learning is to transfer learned knowledge to related situations. 
For example, we may want to control multiple robot arms with slightly different specifications (e.g., link weights or lengths) or different operating environments (e.g., underwater, in low gravity). Normally, learned controllers deal with a single task. In a robotics context, solutions for multiple related tasks are often desired, e.g., for grasping multiple objects~\cite{Kroemer2010} or in robot games, such as robot table tennis~\cite{Mulling2013} or soccer~\cite{Barrett2010}.
Much of the literature on meta and transfer learning in RL has focused on multi-task learning, i.e., cases where the system\slash robot is the same, but the task changes~\cite{Ijspeert2002a,Taylor2007,Barrett2010,daSilva2012,Kober2011a,Konidaris2012,Mulling2013,Deisenroth2014a,finn2017model}. Although meta learning given multiple or non-stationary dynamics has also been considered in~\cite{Velez2016,Killian2017,Cully2015,Maruan2018}.

We adopt a meta learning~\cite{schaul2010metalearning, vilalta2002perspective, finn2017model} perspective on the problem of using knowledge from prior tasks for more efficient learning of new ones. We take a probabilistic view and propose to transfer knowledge within a model-based RL setting using a latent variable model. We focus on settings where system specifications differ, but where the task objective is identical. We treat system specifications as a latent variable, and infer these unobserved factors and their effects online. To address the issue of meta learning within the context of data-efficient RL, we propose to learn predictive dynamics models conditioned on the latent variable and to learn controllers using these models. We use Gaussian processes (GPs)~\cite{Rasmussen2006} to model the dynamics, and MPC for policy learning. To obtain a posterior distribution on the latent variable, we use variational inference. The posterior can be updated online as we observe more and more data, e.g., during the execution of a control strategy. Hence, we systematically combine three orthogonal ideas (probabilistic models, MPC, meta learning) for increased data efficiency in settings where we need to solve different, but related tasks.

\section{MODEL-BASED RL}
We consider stochastic systems of the form
\begin{align}
\vec x_{t+1} = f(\vec x_t, \vec c_t) + \vec \epsilon
\label{eq:dynamical_system}
\end{align}
with state variables $\vec x \in \mathbb{R}^D$, control signals $\vec c \in \mathbb{R}^K$ and i.i.d. system noise $\vec \epsilon \sim \mathcal N(\vec 0, \mat E)$, where $\mat E = \textnormal{diag}(\sigma^2_1,\dotsc,\sigma^2_D)$. For model-based RL we first aim to learn the unknown transition function $f$.
In this context, \cite{Schneider1997,Deisenroth2011c} highlighted that probabilistic models of $f$ are essential for data-efficient learning as they mitigate the effect of model errors. Therefore, we learn the dynamics of the system using a GP.

\paragraph{GP Dynamics}
A GP is a probabilistic, non-parametric model and can be interpreted as a distribution over functions. A GP is defined as an infinite collection of random variables $\{f_1, f_2, \dotsc\}$, any finite number of which are jointly Gaussian distributed~\cite{Rasmussen2006}. A GP is fully specified by a mean function $m$ and a covariance function (kernel) $k$, which allows us to encode high-level structural assumptions on the underlying function such as smoothness or periodicity. We denote an unknown function $f$ that is modeled by a GP by $f\sim GP(m(\cdot), k(\cdot, \cdot))$. We use the squared exponential (RBF) covariance function
\begin{align}\label{eq:covariance}
\hspace{-2.5mm}k(\x_i, \x_j)\! =\! \sigma_f^2\exp\big(-\tfrac{1}{2}(\x_i - \x_j)^T \mathbf{L}^{-1} (\x_i - \x_j)\big)
\end{align}
where $\sigma_f^2$ is the signal variance and $\mat L$ is a diagonal matrix of squared length-scales.

\paragraph{RL with MPC}
Our objective is to find a sequence of optimal controls $\vec c_0^*,\dotsc,\vec c_{H-1}^*$ that minimizes the expected finite-horizon cost
\begin{align}
J = \E\left[\sum\nolimits_{t = 1}^H \ell(\vec x_t) \right],
\label{eq:J function}
\end{align}
where $\vec x_t$ is the state of the system at time $t$ and $\ell$ is a known immediate\slash instantaneous cost function that encodes the task objective. We consider an episodic setting. Initial states $\vec x_0$ are sampled from
$p(\vec x_0)=\mathcal N(\vec\mu_0,\mat\Sigma_0)$.

To find the optimal open-loop sequence $\vec c_0^*, \dotsc, \vec c_{H-1}^*$, we compute the expected long-term cost $J$ in~\eqref{eq:J function} using Gaussian approximations $p(\vec x_1), \dotsc, p(\vec x_H)$ for a given control sequence $\vec c_0, \dotsc, \vec c_{H-1}$. The computation of the expected long-term cost is detailed in the supplementary material. Then, we find an open-loop control sequence that minimizes the expected long-term cost and apply the first control signal $\vec c_0^*$ to the system, which transitions into the next state. Next we re-plan, i.e., we determine the next open-loop control sequence $\vec c_0^*, \dotsc, \vec c_{H-1}^*$ from the new state. This iterative MPC approach turns an open-loop controller into a closed-loop controller. Combining MPC with learned GP models for the underlying dynamics increases the robustness to model errors and has shown improved data efficiency in RL~\cite{Kamthe2018}.

\section{MODEL-BASED META RL}
\label{sec:meta_learning}
We assume a setting with a potentially infinite number of dynamical systems that are of the same type but with different specifications (e.g., multiple robotic arms with links of differing lengths and weight). More formally, we assume a distribution over dynamical systems with samples $f_p \sim p(f)$ indexed by $p=1..P$. Each sample $f_p$ is a dynamical system of the form 
\eqref{eq:dynamical_system} with states $\vec x \in \mathbb{R}^D$ and control signals $\vec c \in \mathbb{R}^K$. Instead of learning individual predictive models for each dynamical system from scratch, we look to meta learning as an approach to learning new dynamics more data efficiently by leveraging shared structure in the dynamics.
\paragraph{Meta Learning}
Generally, meta learning aims to learn new tasks with minimal data and/or computation using knowledge or inductive biases learned from prior tasks \cite{finn2017model}. Here we require our model to accomplish two things simultaneously: 
\begin{enumerate}
\item \textbf{Multi-Task Learning}: Disentangle global and task-specific properties of the different dynamics such that it can solve multiple tasks.
\item \textbf{Transfer Learning}: Use global properties to generalize predictive performance to novel dynamics.
\end{enumerate}
We propose to address this meta-learning challenge in a probabilistic way: We model the distribution over systems using a latent embedding $\vec h$ and model the dynamics using a global function conditioned on the latent embedding. Each sample $f_p$ from the distribution is modeled as
\begin{align}
\vec x_{t+1} = f(\vec x_t, \vec c_t, \vec h_p) + \vec \epsilon\,,
\label{eq:multi_dynamical_system}
\end{align}
such that the successor state depends on the latent system specification $\vec h_p$. This means, we explicitly model the global properties through a shared function $f$ and the task-specific variation using a distribution over the latent variables $p(\vec h_p)$. Framing the meta learning problem as a hierarchical Bayesian model means that meta-training becomes inference in a meta-learning model.
\paragraph{Training and Evaluation}
Training corresponds only to the \emph{multi-task learning} aspect of our meta learning approach. We aim to learn the global function $f$ and the latent embeddings $\vec h_p$ given trajectory observations from a set of training systems. For evaluation at test time, we use inference to obtain a distribution over a set of latent variables $\vec h_*$ for each test system. Since our objective is to improve data efficiency in an RL setting, we consider two related but distinct measures of performance. One corresponds to the transfer learning aspect of our approach, where we infer only the latent test embeddings without updating the global model $f$. We refer to this as the \emph{single-shot performance}. The other measure we use is the additional data required to successfully solve a RL task: \emph{few-shot learning}. In this case, the global model $f$ is updated with new additional data, thus combining both the multi-task and transfer learning aspects.

The meta RL procedures for training and testing are detailed in algorithms~\ref{alg:meta_training} and~\ref{alg:meta_test},	 respectively.
\begin{algorithm}[h]
 \label{alg:meta_training}
 Initialize dataset $D$ and model $M$\\
 \Comment{Initial random rollouts}\\
 \ForAll{training tasks}
  	{execute random policy\\
    add observations to $D$}
 \Comment{Meta training}\\
 \While{training tasks not solved}{
 |\textbf{update}|: train $M$ and infer $\vec{h}$ given $D$\\
  \ForAll{unsolved training tasks}
  	{\For{each step in horizon}{
    	|\textbf{plan}|: get control sequence using~\eqref{eq:J function}\\
        |\textbf{execute}|: execute first control in sequence
    	}
    add observations to $D$\\
    check if task solved}
    }
 \caption{Model-based Meta RL with MPC (Train)}
\end{algorithm}
\begin{algorithm}[h]
 \label{alg:meta_test}
 Given dataset $D$ and model $M$ from training\\
 \Comment{Single shot performance}\\
 \ForAll{test tasks}
  	{\For{each step in horizon}{
    	|\textbf{plan}|: get control sequence using~\eqref{eq:J function}\\
        |\textbf{execute}|: execute first control in sequence\\
        |\textbf{inference}|: infer the value of $\vec{h}_*$ given observations so far
    	}
    add observations to $D$\\
    check if task solved}
 \Comment{Meta test}\\
 \While{test tasks not solved}{
 |\textbf{update}|: train $M$ and infer $\vec{h}$ given $D$\\
  \ForAll{unsolved test tasks}
  	{\For{each step in horizon}{
    	|\textbf{plan}|: get control sequence using~\eqref{eq:J function}\\
        |\textbf{execute}|: execute first control in sequence
    	}
    add observations to $D$\\
    check if task solved}
    }
 \caption{Model-based Meta RL with MPC (Test)}
\end{algorithm}
\subsection{META-LEARNING MODEL}  
Our meta-learning model is a GP prior on the unknown transition function in~\eqref{eq:multi_dynamical_system} with a concatenated state $\tx_t = (\x_t, \cs_t, \h_p) \in \mathbb{R}^{D+K+Q}$ as the input to the model. We define $\vec y_t = \vec x_{t+1} - \vec x_t$ as the targets of the GP and take the mean function to be $m(\tilde{\vec x}_t) = \vec 0$, which encodes that a priori the state does not change~\cite{Deisenroth2015}. Each dimension of the targets $\vec y$ is modeled by an independent GP.  We use a Gaussian likelihood
\begin{align}
\label{eq:likelihood}
p(\y_t|\tx_t, \f(\cdot), \vec \theta) = \mathcal{N}(\y_t|\f(\tx_t), \mat E),
\end{align}
where $\vec \theta=\{\mat E, \mat L, \sigma^2_f, Q\}$ are the model hyper\-parameters and  $\f(\cdot) = \big(f^1(\cdot),\dotsc,f^D(\cdot)\big)$ denotes a multi-dimensional function. We place a standard-normal prior $\h_p\sim\mathcal{N}(\boldsymbol{0}, \I)$ on the latent variables $\vec h_p$. The full specification of the model is
\begin{align}
\label{eq:model}
&p(\mat Y, \mat H, \f(\cdot)|\mat X, \mat C) \\ 
&=\prod\nolimits_{p=1}^P p(\vec h_p)\prod\nolimits_{t=1}^{T_p}
p(\y_t|\vec x_t, \vec c_t, &\vec h_p, \f(\cdot))p(\f(\cdot))\nonumber
\end{align}
where we denote a collection of vectors in bold upper-case and we have dropped dependence on the hyperparameters for notation purposes. The corresponding graphical model is given in Fig.~\ref{fig:graphical_model}. The figure shows the dependence of individual system observations on the global GPs $\vec f(\cdot)$ modeling each dimension of the outputs, the system-specific latent embeddings $\vec h_p$ and the observed states and controls.
\paragraph{Model Properties} Our meta-learning GP ($\modelname$) model exhibits three important properties:
\begin{enumerate}
\item The latent variable encodes a distribution over plausible systems and is inferred from data
\item Conditioning the GP on the latent variable enables it to disentangle global and task specific variation in the dynamics. Generalization to new dynamics is done by inferring the latent variable of that system.
\item The latent variable is fixed within system trajectories so that inference can be performed online (e.g. while executing a controller).
\end{enumerate}
Fig.~\ref{fig:lvgpd_properties} illustrates these properties on a toy example.
\begin{figure}[h]
  \begin{center}
    \begin{tikzpicture}[x=1.5cm,y=0.5cm, scale=0.5]

	\node[obs] (yt) {$\vec{y}_t$};
    \node[latent, above=of yt] (f) {$\overset{\mathrm{\infty}}{\vec{f}(\cdot)}$};
    \node[const, left=of yt] (xt) {$\vec{x}_t$};
    \node[const, below=of xt] (ct) {$\vec{u}_t$};
    \node[latent, right=of yt] (hp) {$\vec{h}_p$};

    \edge {f} {yt};
    \edge {xt} {yt};
    \edge {ct} {yt};
    \edge {hp} {yt};
    
    
    \plate {traj} {
    (xt)
    (ct)
    (yt)
    } {$t=1,\dotsc, T_p$};
    
	\plate {} {
    (traj)
    (hp)
    } {$p=1,\dotsc, P$};

\end{tikzpicture}



    
    
    

  \end{center}
  \caption{Graphical model for our $\modelname$ model.
  }
  \label{fig:graphical_model}
  \figspace
\end{figure}
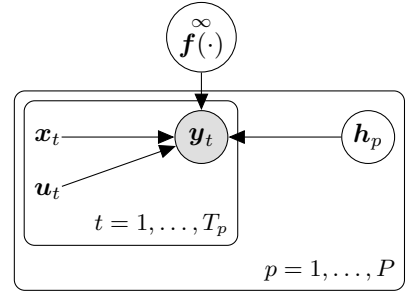
\begin{figure}[htb]
\centering
\subfloat{%
  \includegraphics[width=1.\linewidth]{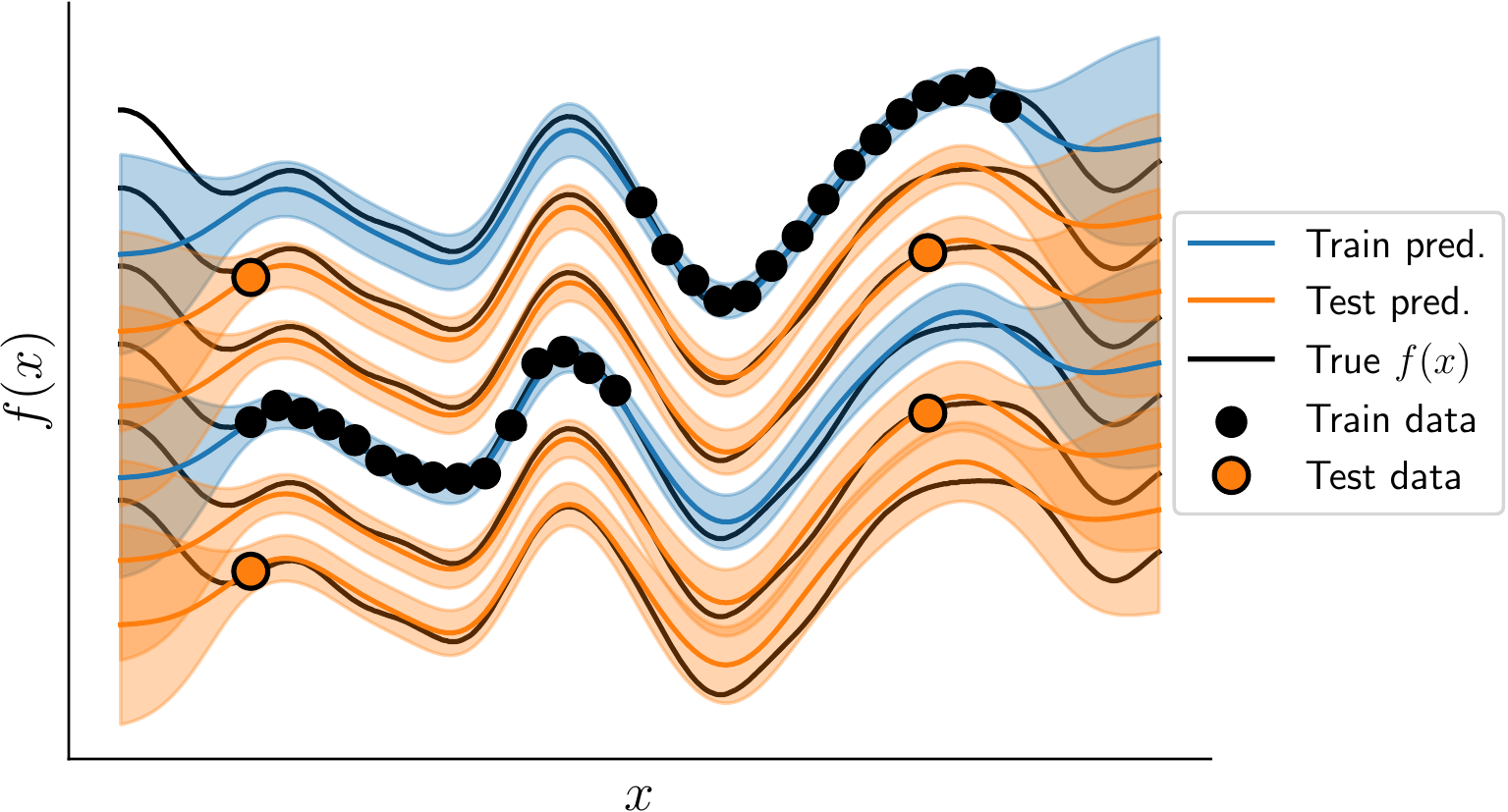}
  \label{fig:pred sgp}
}
\caption{The figure shows six unknown tasks (toy examples) with a shared structure (the same function) and task specific variation (fixed offset). The $\modelname$ model is able to disentangle the two automatically given the training data (black discs) as demonstrated by the training prediction curves. It also infers a reasonable value for the offset given a single observations from unseen test tasks (orange discs) and can use the global structure to generalize predictive performance on those tasks.}
\label{fig:lvgpd_properties}
\figspace
\end{figure}
\subsection{INFERENCE}
\label{sec:inference}
To learn the dynamics model we seek to optimize the hyperparameters $\vec \theta$ w.r.t. the log-marginal likelihood, which involves marginalization of the latent variables in~\eqref{eq:model}. For predictions of the evolution of a system we also need to infer the posterior GP and the posterior distribution of the latent variables $\hall = (\vec h_1,...,\vec h_P)$. We approach this problem with approximate variational inference. We posit a variational distribution that assumes independence between the latent functions of the GP and the latent task variables
\begin{align}
Q(\f(\cdot), \mat H) = q(\f(\cdot))q(\mat H)
\label{eq:approximate posterior}
\end{align}
and minimize the Kullback-Leibler divergence between the approximate and true posterior distributions. Equivalently we can maximize the evidence lower bound
\begin{align}
\mathcal{L} = \mathbb{E}_{Q(\f(\cdot), \mat H)}\Big[\log\frac{p(\mat Y, \mat H, \f(\cdot)|\mat X, \mat C)}{Q(\f(\cdot), \mat H)}\Big],
\label{eq:lower bound 1}
\end{align}
which lower-bounds the log-marginal likelihood~\cite{Hoffmann2013}.
We parameterize our variational distribution such that we can compute the lower bound in \eqref{eq:lower bound 1}. We then jointly optimize $\mathcal{L}$ with respect to the model hyperparameters and the variational parameters. 

\paragraph{Sparse Gaussian Processes}
It is important to account for the fact that training a GP on a joint data set of $P$ different systems quickly becomes infeasible due to the $\mathcal{O}(T^3)$ computational complexity for training and $\mathcal{O}(T^2)$ for predictions where $T$ is the total number of observations. To address this we turn to the variational sparse GP approximation \cite{Titsias2009VariationalProcesses} and approximate the posterior GP with a variational distribution $q(\f(\cdot))$ that depends on a small set of $M \ll T$  \emph{inducing points}. We introduce a set of $M$ inducing inputs $\Zi = (\zi_1,\dotsc,\zi_M) \in \mathbb{R}^{M\times(D+K+Q)}$, which live in the same space as $\tx$, with corresponding GP function values $\Ui = (\ui_1,\dotsc,\ui_M) \in \mathbb{R}^{M\times D}$. We follow \cite{Hensman2013} and specify the variational approximation as a combination of the conditional GP prior and a variational distribution over the inducing function values, independent across output dimensions
\begin{align}
\label{eq:var_u}
q(f^d(\cdot)) = \int p(f^d(\cdot)|\ui^d)q(\ui^d)d\ui^d.
\end{align}
where $q(\ui^d) = \mathcal{N}(\ui^d| \m^d, \Sc^d)$ is a full rank Gaussian. The integral in (\ref{eq:var_u}) can be computed in closed form since both terms are Gaussian, resulting in a GP with mean and covariance functions given by
\begin{align}
\label{eq:post_mean}
m_q(\cdot) &= \vec k_Z^T(\cdot)\mat K_{ZZ}^{-1}\vec m^d \\
\label{eq:post_var}
k_q(\cdot, \cdot) &= k(\cdot, \cdot) - \vec k_Z^T(\cdot)\mat K_{ZZ}^{-1}(\mat K_{ZZ} - \mat S^d)\mat K_{ZZ}^{-1}\vec k_Z(\cdot)
\end{align}
where $[\vec k_Z(\cdot)]_i = k(\cdot, \zi_i)$ and $[\mat K_{ZZ}]_{ij} = k(\zi_i, \zi_j)$. Here, the variational approach has two main benefits: a) it reduces the complexity of training to $\mathcal{O}(TM^2)$ and predictions to $\mathcal{O}(TM)$, b) it enables mini-batch training for further improvement in computational efficiency. 

\paragraph{Latent Variables}
For the latent variables $\hall$ we assume a Gaussian variational posterior
\begin{align}\label{eq:fullpost}
q(\hall) = \prod\nolimits_{p=1}^P \mathcal{N}(\h_p|\n_p, \Tc_p)
\end{align}
where $\Tc_p$ is in general a full rank covariance matrix. We use a diagonal covariance in practice for more efficient computation of the ELBO \eqref{eq:lower bound 1}.
\paragraph{Evidence Lower Bound (ELBO)}
The ELBO can be shown to decompose into (see supplementary material)
\begin{align}
\mathcal{L} &= \sum\nolimits_{p=1}^P\sum\nolimits_{t=1}^{T} \mathbb{E}_{q(\f_t|\x_t, \cs_t)}\big[\log p(\y_t|\f_t)\big]\nonumber\\ 
 &\quad-\!\KL\big[q(\hall)||p(\hall)\big] \!-\! \KL\big[q(\Ui)||p(\Ui)\big]
 \label{eq:lbound}
\end{align}
where the expectation is taken with respect to
\begin{align}\label{eq:qf}
q(\f_t|\x_t, \cs_t) = \int q(\f_t|\x_t, \cs_t, \h_p)q(\h_p)d\h_p.
\end{align}
We emphasize that $q(\f_t|\x_t, \cs_t, \h_p) = q(\f(\tx_t)|\x_t, \cs_t, \h_p)$ is the marginal of the GP evaluated at the inputs $\tx_t$. The integral in \eqref{eq:qf} is intractable due to the non-linear dependence on $\h_p$ in~\eqref{eq:post_mean} and \eqref{eq:post_var}. Given our choice of kernel (RBF) and Gaussian variational distribution $q(\h_p)$ the first and second moments can be computed in closed form. We could use these terms to compute the log-likelihood term in closed form since the likelihood is Gaussian but in practice this can be prohibitively expensive since it requires the evaluation of a $TM^2D$ tensor. Instead we avoid computing the moments by approximately integrating out the latent variable using Monte Carlo sampling.

\paragraph{Training} For the update steps in algorithms \ref{alg:meta_training} and \ref{alg:meta_test} we jointly optimize the GP hyperparameters $\vec \theta$ and the variational parameters $\boldsymbol{\phi}=\{\Zi, M\, \{\m^d, \Sc^d\}_{d=1}^D, \{\n_p, \Tc_p\}_{p=1}^P\}$ w.r.t. the ELBO. For the inference step in algorithm \ref{alg:meta_test}, we optimize only the variational parameters for the latent variables $\vec{h}$, i.e. $\boldsymbol{\phi}_{\vec{h}}=\{\n_p, \Tc_p\}_{p=1}^P$.

In practice, we use a single sample $\h_p \sim q(\h_p)$ drawn from the variational distribution for each system. We use stochastic mini-batch training, sampling a small number of trajectories and their associated latent variable at a time. Empirically, we found standardizing the input states and controls ($\vec{x}, \vec{c}$) and outputs ($\vec{y}$) crucial for successful training of the model. For optimization we used Adam \cite{KingmaB14} with default hyperparameters: $\alpha = 1\times 10^{-2}, \beta_1 = 0.9, \beta_2 = 0.999, \epsilon = 10^{-8}$.
\section{EXPERIMENTS}
Our experiments focus on evaluating our proposed model in terms of predictive performance, the nature of the latent embeddings and data efficiency. We address the following questions:
``Does conditioning the GP on the latent variable allow us to disentangle system specific and global properties of the observations? Does this improve predictive performance in the transfer learning setting?'' (Section~\ref{subsec:res-pred}).
``Is the latent system embedding the model learns a sensible one? (Section~\ref{subsec:res-latent}) ``Does the application of our $\modelname$ in model-based RL lead to data-efficient learning across tasks'' (Section~\ref{subsec:res-rl}).

As a baseline model we use a sparse GP (SGP)~\cite{Titsias2009VariationalProcesses} as described in Section~\ref{sec:inference} but without the latent variable that explicitly represents the task. For assessing the model quality (Section~\ref{subsec:res-pred}) we additionally evaluate the performance of a standard GP with no sparse approximation. We use the following nonlinear dynamical systems to perform our experiments:
\paragraph{Cart-pole swing-up} The cart-pole system consists of a cart that moves horizontally on a track with a freely swinging pendulum attached to it. The state of this nonlinear system is the position $x$ and velocity $\dot x$ of the cart and the angle $\theta$ and angular velocity $\dot\theta$ of the pendulum. The control signals act as a horizontal force on the cart limited to the range $c \in [-15, 15]\,\mathrm{N}$. The mean of the initial state distribution is the state where the pendulum is hanging downward. The task is to learn to swing up and balance the pendulum in the inverted position in the middle of the track.

\paragraph{Double-pendulum swing-up} The double-pendulum system is a two-link robotic arm with two motors, one in the shoulder and one in the elbow. The state of the system comprises the angles $\theta_1,\theta_2$ and angular velocities $\dot\theta_1,\dot\theta_2$ of the inner and outer pendulums, respectively. The control signals are the torques  $c_{1,2}\in [-4,4]\,\mathrm{Nm}$ applied to the two motors. The mean of $p(\vec x_0)$ is the position where both pendulums are hanging downward. The goal is to find a control strategy that swings the double pendulum up and balances it in the inverted position.

\subsection{QUALITY OF MODEL LEARNING}
\label{subsec:res-pred}
In the first set of experiments, we investigate if the latent variable of the $\modelname$ improves prediction performance on unseen systems compared to the $\baseline$ baseline. To assess the effect of the sparse approximation we also include a standard GP baseline (no sparse approximation) in this section.
To test the prediction quality, we execute the same fixed control signals\footnote{the control signals were manually chosen as ones that solved a configuration not included in either the training or test set.} on six settings of the cart-pole task to generate one $100$-step ($10\,\mathrm{s}$) trajectory per training task. The specifications of the training tasks were all combinations $(m,l)$ of $m \in \{0.4, 0.6, 0.8\},~l \in \{0.5, 0.7\}$ where $m$ and $l$ denote the mass and length of the pendulum, respectively. Thus,  the total number of data points for our six training tasks is $T=600$ amounting to $60\,\mathrm{s}$ of interaction time.

For evaluation we use the same sequence of control signals we used for training and compute the one-step prediction quality in terms of root mean squared error (RMSE) and negative log likelihood (NLL) on a set of test tasks. We use $14$ held-out test tasks specified as $m \in \{0.4, 0.6, 0.7, 0.8, 0.9\},~l \in \{0.4, 0.5, 0.6, 0.7\}$, excluding the $(m,l)$-combinations  of the training tasks.

During evaluation, we observe $10$ time steps from an unseen trajectory based on which we infer the latent task $\vec h_p$ using variational inference for the $\modelname$ while leaving the model hyperparameters and other variational parameters fixed. We then predict the next $90$ steps using the $\modelname$, $\baseline$ and GP models. $\modelname$  also performs online inference of the latent variable after each step. We repeat this experiment with $10$ different seeds that determine the initial state, and average the results.

\begin{figure}[h]
\centering
\includegraphics[width=0.95\linewidth]{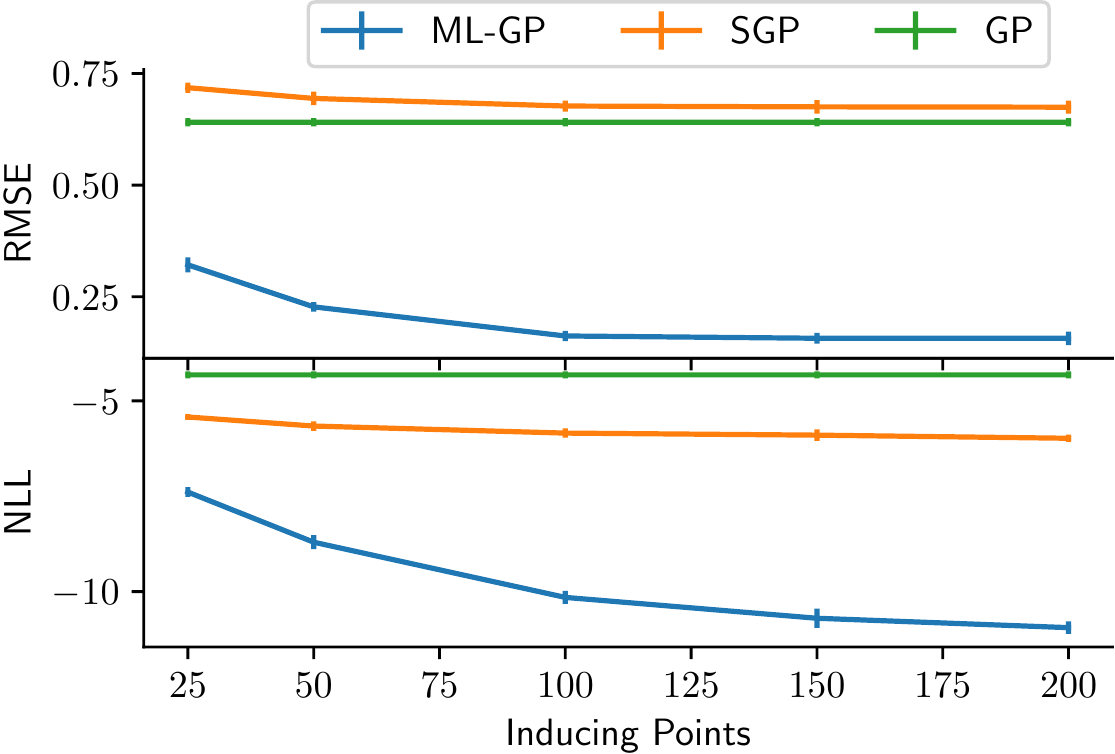}
\caption{Mean and two standard deviation confidence error-bars of the RMSE and NLL for the $\modelname$, $\baseline$ and the standard GP model as a function of the number of inducing points. The $\modelname$ significantly outperforms both baselines.}
\label{fig:modelquality}
\figspace
\end{figure}
\begin{figure}[h]
\centering
\includegraphics[width=0.95\linewidth]{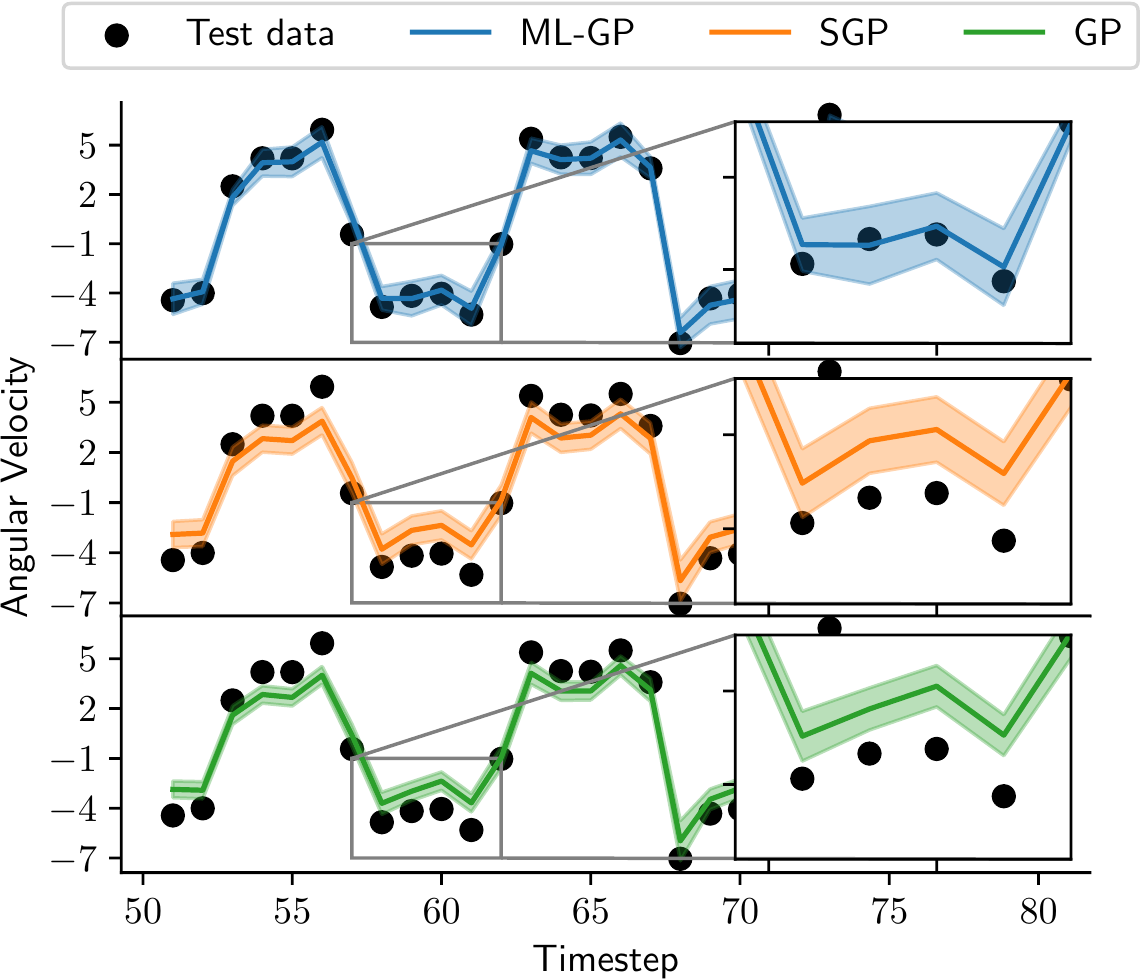}
\caption{One-step predictions of the angular velocity in cart-pole.
The figure shows the true data points (discs) and the predictive distributions with a two standard deviation confidence interval for the $\modelname$, $\baseline$ and a standard GP. The $\modelname$ generalizes well to new tasks; both the $\baseline$ and GP baselines are overly confident.}
\label{fig:overconfidence}
\figspace
\end{figure}
Fig.~\ref{fig:modelquality} shows the RMSE and NLL for all 3 models. The $\modelname$ clearly outperforms both the $\baseline$ and GP baselines in terms of both the accuracy of its mean predictions (as evident by the RMSE) as well as capturing the data better under its predictive distribution as seen by the NLL. The NLL accounts for both the mean prediction as well as the uncertainty of the model about the prediction. Both baselines have comparable RMSEs to each other with enough inducing points but generalize poorly on new tasks with overconfident predictions. Fig.~\ref{fig:overconfidence} illustrates this behavior.

The baselines fail to generalize since they have no observations from the system with this configuration. The $\modelname$  generalizes from training to new test tasks naturally because it explicitly incorporates the latent variables encoding the system configuration.

\subsection{LATENT EMBEDDING}
\label{subsec:res-latent}
In order for our model to perform well in meta learning settings, the latent variables $\vec h_p$ need to reflect a \textit{sensible} embedding. By sensible we mean it should take on a particular structure: a) locally similar values in the latent space should correspond to similar task specifications and b) moving in latent space should correspond to coherent transitions in task specifications. 

\begin{figure}[h]
\centering
\includegraphics[width=0.95\linewidth]{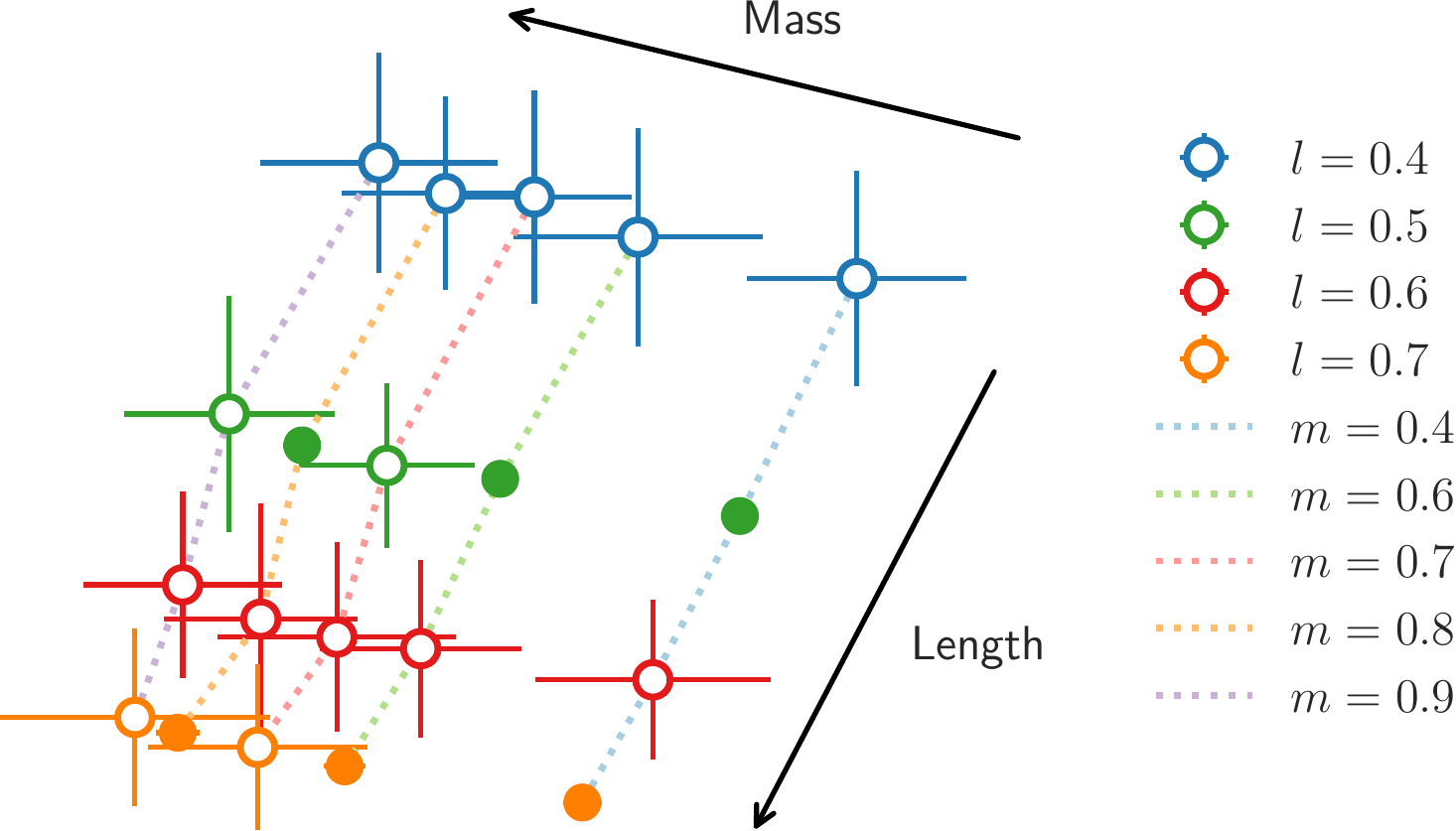}
\caption{Latent space embedding of cart-pole configurations\slash tasks. The figure shows the mean (discs) of the inferred latent variables and two standard deviation error bars. Filled discs are training tasks and empty discs are held out test tasks. The colors of the discs represent the length and the colors of the dotted lines between discs represent the mass.}
\label{fig:latent_cartpole}
\figspace
\end{figure}
Fig.~\ref{fig:latent_cartpole} shows an example of an inferred latent embedding of both training and test tasks after the training procedure outlined above. The test-task latent variables are inferred from 10 observations from the held-out systems.

The different colors of the discs denote the four different settings of lengths whereas the colors of the dotted lines connecting the discs denote the five different settings of mass. The figure plots the mean of each $q(\vec h_p)$ with two standard deviation error bars in each dimension. The embedding displays an intuitive structure where changes in length or mass are disentangled (denoted by the black arrows) into a length-mass coordinate system with the expected transitive properties, e.g. the lengths are ordered as blue ($l=0.4$), green ($l=0.5$), red ($l=0.6$) and orange ($l=0.7$). The uncertainty estimates also exhibit qualitatively the intuitive property of being less uncertain about tasks which are similar to (closer to) the training tasks, e.g. comparing the red and blue tasks in fig.~\ref{fig:latent_cartpole}.  

\subsection{DATA-EFFICIENT RL}
\label{subsec:res-rl}

Our second set of experiments investigates the performance of the $\modelname$ model in terms of data efficiency in RL settings. Specifically, we look at whether our meta learning approach is a) at least as efficient at solving a set of training tasks, b) more efficient at solving subsequent test tasks, when compared to a non-meta learning baseline and c) whether the $\modelname$ model improves performance when compared to the $\baseline$ model trained with the meta learning approach.

We first learn a model of the dynamics~\eqref{eq:multi_dynamical_system}, which we then use to learn a policy to control the system. For policy learning we use MPC, minimizing the cost in~\eqref{eq:J function} with a moving horizon to learn an optimal sequence of control signals. We assume we have a set of training systems and evaluate the performance of the models using some held-out test systems with novel configurations (tasks).

We run experiments on both the cart-pole swing-up task and the double-pendulum swing-up task. In both scenarios, we use a sampling frequency of $10\,\mathrm{Hz}$, episodes of $30$ steps ($3\,\mathrm{s}$) and a planning horizon of $10$ steps. For the cart-pole swing-up, solving the task means the pendulum is balanced closer than $8\,\mathrm{cm}$ from the goal position for at least the last $10$ steps. For the double-pendulum swing-up, it means the outer pendulum is balanced closer than $22\,\mathrm{cm}$ for at least the last $10$ steps.

At meta-training or test-time, a pass through the training-/test-set means executing the MPC policy learning algorithm on each of the unsolved task in that set. Each execution constitutes a trial for that task. The sets are traversed until all the tasks are solved or all unsolved tasks have executed 15 trials. The training and test procedures are detailed in algorithms \ref{alg:meta_training} and \ref{alg:meta_test} in section \ref{sec:meta_learning}. All results are averaged over 20 independent random initializations.

Note that we execute on all (unsolved) tasks before retraining the dynamics model as detailed in section~\ref{sec:meta_learning}. This means that the model is updated with $3\,\mathrm{s}$ worth of experience for every task in that pass at a time. On the other hand, the model does not take advantage of additional prior experience until it has completed a pass. 

For comparison with the $\modelname$ model, we use the $\baseline$ model trained in two different ways. To establish a lower-bound baseline, we run the model-based RL approach where we train a separate model for each task on both the training and test sets. After each training task we additionally attempt to solve each of the test tasks to evaluate single-shot performance where we report the mean across the training tasks as the single shot success rate. We refer to this baseline as $\baseline$-I which is a sparse variant of the approach in \cite{Kamthe2018} that achieves state-of-the-art in data efficiency. Secondly, we train a single $\baseline$ model on all the training tasks simultaneously using the same training approach as we do for $\modelname$. We refer to this baseline as $\baseline$-ML.
\paragraph{Cart-pole swing-up} We train the models on six specifications of the cart-pole dynamics, with $m \in\{0.4, 0.6, 0.8\},~l \in \{0.6, 0.8\}$ and evaluate its performance on a set of four test tasks chosen as $m \in\{0.7, 0.9\},~l = \{0.5, 0.7\}$. We choose these settings to examine the performance on both interpolation and extrapolation for differing lengths and masses. We choose the squared distance between the tip of the pendulum and goal position (with the pendulum balanced straight in the middle of the track) as the cost.
\begin{figure}[ht]
\centering
\includegraphics[width=0.95\linewidth]{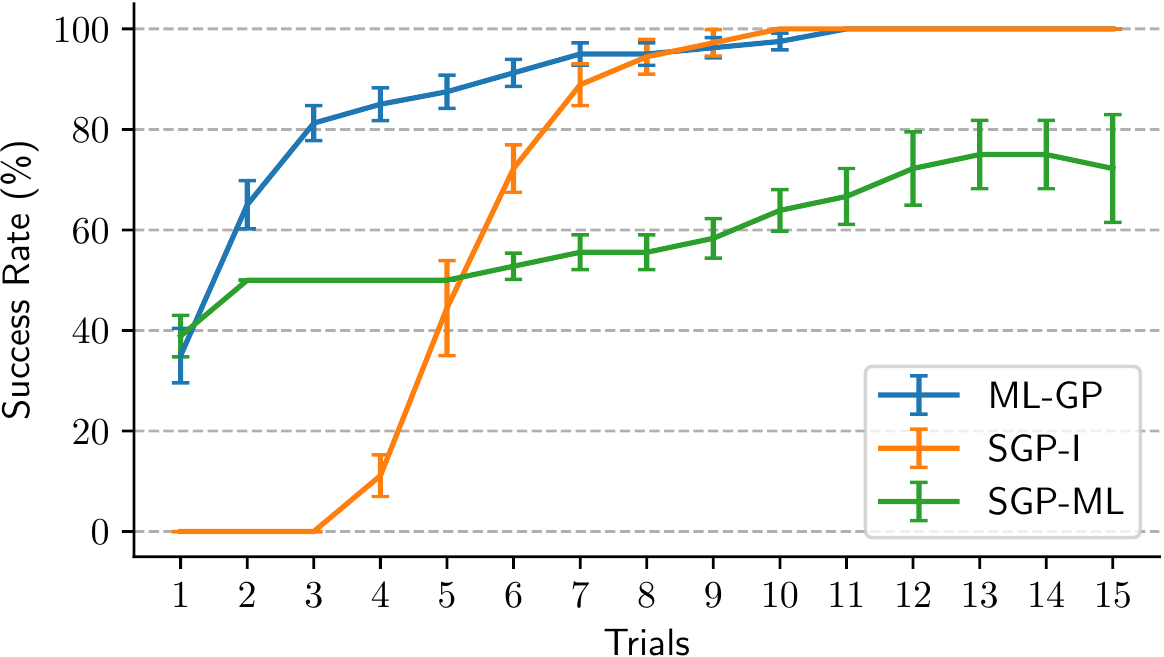}
\caption{Mean success rate over initializations and the four test tasks for the cart-pole system after training on six tasks. The graph compares $\modelname$ with $\baseline$-I (trained independently) and $\baseline$-ML (trained on all tasks).}
\label{fig:cartpole_success}
\figspace
\end{figure}
Fig.~\ref{fig:cartpole_success} shows the mean success rate (over initializations and the four test tasks) of $\modelname$, $\baseline$-I and $\baseline$-ML against the number of trials executed on the systems. We observe that both the $\modelname$ model and the $\baseline$-ML display generalization to new tasks as evident by the success rate in the first trial (see also Table~\ref{tab:cartpole-control}). 
However, whereas the $\modelname$ quickly improves with more observations in subsequent trials, the $\baseline$-ML model struggles to solve the remaining tasks. We attribute this failure to the inability of the $\baseline$-ML model to explain variation in the dynamics caused by differences in system specifications.

\begin{figure}[htb]
\centering
\includegraphics[width=0.95\linewidth]{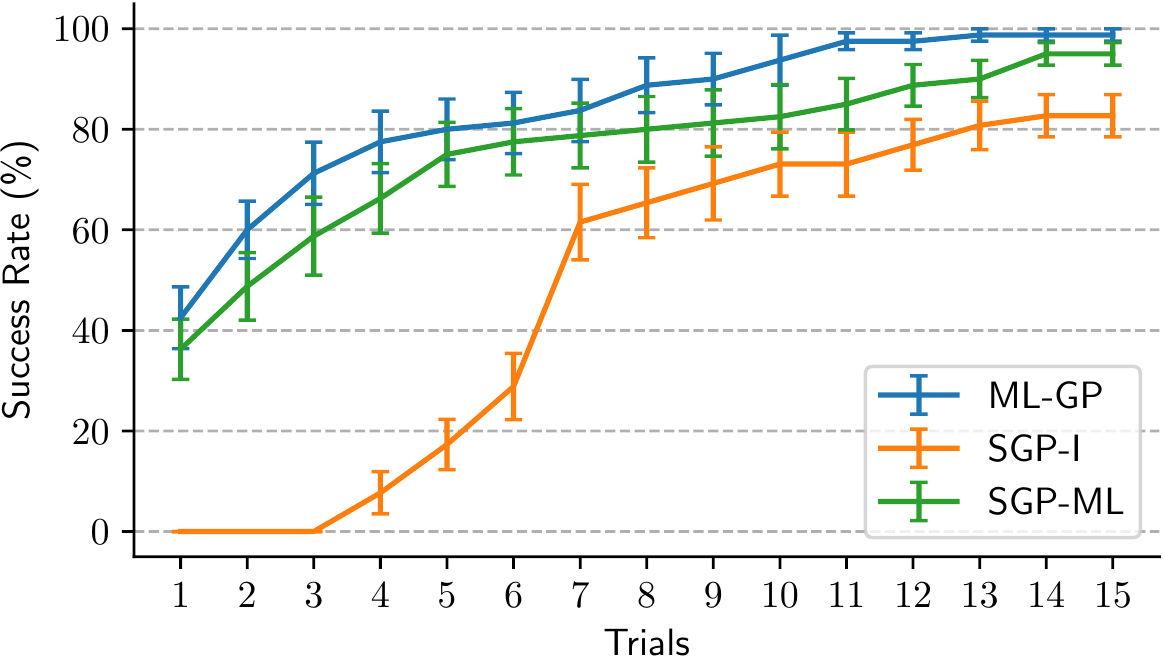}
\caption{Mean success rate over initializations and the four test tasks for the double pendulum  after training on six tasks. The graph compares the $\modelname$ against the $\baseline$-I (trained independently on each task) and the $\baseline$-ML (trained using the meta learning procedure).}
\label{fig:dp_success}
\figspace
\end{figure}

When comparing with independent training of each system we see that the $\modelname$ compares favorably, reaching $80\%$ success rate after only three trials and $90\%$ after six trials compared to the $\baseline$-I, which reaches $80\%$ after $7$ trials and $90\%$ after $8$ trials. We further analyze performance of $\modelname$ to identify the tasks that were additionally solved between trials $3$ and $6$. We find that this is due to a consistently challenging system with $m=0.9, l=0.5$, which requires the learner to extrapolate beyond the range of values seen during training. The mean number of trials required to solve this task is $4.3 \pm 0.6$, compared to the task mean of $2.7 \pm 0.2$ trials.
\begin{table}[htb]
\caption{Mean time spent solving the cart-pole system and the single-shot success rate.}
\label{tab:cartpole-control}
\begin{center}
\scalebox{0.9}{
\begin{tabular}{@{}lccc@{}}
\multicolumn{1}{c}{\bf MODEL} & \multicolumn{1}{c}{\bf TRAIN (s)} & \multicolumn{1}{c}{\bf TEST (s)} & \multicolumn{1}{c}{\bf SINGLE SHOT}  \\
\toprule
$\baseline$-I   & 16.1 $\pm$ 0.4 & 17.5 $\pm$ 0.4 & 0.08 $\pm$ 0.01 \\
$\baseline$-ML    & 23.7 $\pm$ 1.4 & 20.8 $\pm$ 1.2 & \best{0.38 $\pm$ 0.04} \\
$\modelname$ & \best{15.1 $\pm$ 0.5} & \best{~8.1 $\pm$ 0.6} & \blue{0.35 $\pm$ 0.05} \\
\end{tabular}
}
\end{center}
\figspace
\end{table}
Table~\ref{tab:cartpole-control} shows the mean total time required to solve the training and test tasks. On average, $\modelname$ needs less than half the amount of time to solve the test tasks compared to individually training on the tasks ($\baseline$-I). We also see an improvement in the total training time, which suggests that $\modelname$ derives some transfer benefit during training despite training on the systems on a concurrent trial basis, i.e. we do not update the model until all systems have executed a given trial. Compared to the $\baseline$-ML, the $\modelname$ model can maintain an accurate model while learning multiple systems and quickly adapts to new dynamics, whereas the performance of $\baseline$-ML stagnates as reflected in the interaction time on both the training and test systems. 
\paragraph{Double-pendulum swing-up} We repeat the same experimental set-up on the double-pendulum task. We trained on six systems with $m_1 \in\{0.5, 0.7\},~l_1 \in \{0.4, 0.5, 0.7\}$ and evaluate on a set of four test tasks chosen as $m_1 \in\{0.6, 0.8\},~l_1 = \{0.6, 0.8\}$, where $m_1,~l_1$ are the mass and length of the inner pendulum. The cost is the squared distance between the tip of the outer pendulum and the goal position (with both pendulums standing straight up). Fig.~\ref{fig:dp_success} plots the mean success rate against the number of trials executed on the system. Comparing the $\modelname$ model to the $\baseline$-ML we observe comparable single-shot performance and a qualitatively similar learning curve for the test tasks. However, the $\modelname$ reaches $90\%$ success rate about four trials before the $\baseline$-ML, around trial nine, i.e. meta learning achieves a significantly higher data efficiency. Compared to independent training of the tasks using $\baseline$-I, the $\modelname$ leads to significantly less (new) training data needed to solve the tasks.
Table~\ref{tab:dp-control} reports the mean total time required to solve the tasks. Compared to the $\baseline$-ML, the performance of the two is similar, although arguably the $\modelname$ compares favorably in terms of average time needed to solve the test tasks. Compared to the $\baseline$-I, we see improvement during training as well as at test time. The average time needed for $\modelname$ to solve the test environments is reduced to around $40\%$ to that of the $\baseline$-I.  
\begin{table}[h]
\caption{Mean time spent solving the double-pendulum system and the single-shot success rate.}
\label{tab:dp-control}
\begin{center}
\scalebox{0.9}{
\begin{tabular}{@{}lccc@{}}
\multicolumn{1}{c}{\bf MODEL} & \multicolumn{1}{c}{\bf TRAIN (s)} & \multicolumn{1}{c}{\bf TEST (s)} & \multicolumn{1}{c}{\bf SINGLE SHOT}  \\
\toprule
$\baseline$-I   & 18.9 $\pm$ 0.7 & 25.9 $\pm$ 1.5 & 0.07 $\pm$ 0.01 \\
$\baseline$-ML    & \blue{17.9 $\pm$ 1.3} & \blue{13.7 $\pm$ 2.2} & \blue{0.36 $\pm$ 0.06} \\
$\modelname$ & \best{16.6 $\pm$ 1.1} & \best{10.2 $\pm$ 1.6} & \best{0.43 $\pm$ 0.06} \\
\end{tabular}
}
\end{center}
\figspace
\end{table}
\section{RELATED WORK}
Meta learning has long been proposed as a form of learning that would allow systems to systematically build up and re-use knowledge across different but related tasks \cite{schaul2010metalearning,vilalta2002perspective}. MAML is a recent promising model free meta learning approach that learns a set of model parameters that are used to rapidly learn novel tasks \cite{finn2017model}. Another interpretation of MAML is formulated in \cite{Grant2018}, which shares our hierarchical Bayesian formulation of the meta learning problem. However, the model-free setting in which MAML has been applied so far typically require orders of magnitude more training data than the model-based approaches we build up on in the present work.

Our $\modelname$ model resembles the GP latent variable model (GPLVM), which is typically used in unsupervised settings~\cite{lawrence2004gaussian}. In the GPLVM, the GP is used to map a low-dimensional latent embedding to higher-dimensional observations. A Bayesian extension (BGPLVM) was introduced in~\cite{titsias2010bayesian} where inference over the latent variable is performed using variational inference. To enable minibatch training, and unlike BGPLVM, we take the approach of~\cite{Hensman2013} and do not marginalize out the inducing variables. The main difference of our model and the GPLVM is that we learn a mapping from both observed and latent inputs to observations.

The combination of observed and latent inputs was investigated in \cite{Wang2012} where the authors use Metropolis sampling for inference which does not scale to larger datasets. A similar setup is found in \cite{Damianou2015} where the model is used for partially observed input data. The work also proposes uses in autoregressive settings similar to ours. Different from us, the distribution over inducing variables is analytically optimized, making minibatch training infeasible.

A related and complimentary line of research are multi-output GPs (MOGPs) \cite{alvarez2012kernels}. Recently, \cite{dai2017efficient} proposed a latent variable extension to MOGPs (LVMOGP) which is similar to our $\modelname$, particularly in their missing data formulation of the model. The crucial difference from our work is that we augment the input space by concatenating the latent variable to the input space while the LVMOGP uses the Kronecker product of two separate kernels applied on the latent and input spaces respectively. Notably, the two models are equivalent for kernels that naturally decompose as a Kronecker product (e.g. the RBF) but depart from there.

A similar framework to ours is found in \cite{Velez2016}, called hidden parameter Markov decision processes (HiP-MDP), which parametrizes a family of related dynamics through a low dimensional latent embedding. The HiP-MDP assumes a fixed latent variable within trajectories. Different from us, the authors use an infinite mixture of GP basis functions where the task specific variation is obtained through the weights of the basis functions \cite{Velez2016}. This work was extended in \cite{Killian2017}, replacing the GP basis functions with a Bayesian neural network. This enables non-linear interactions between the latent and addresses scalability. In this work, the interactions between latent and state variables are obtained through the non-linear RBF kernel, and the scalability is addressed through the variational sparse approach.

In \cite{Deisenroth2014a}, an RL setting is considered that is closely related to our meta-learning set-up. The authors use a parametric policy that depends on a known deterministic task variable and augment the policy function to include it as well. In~\cite{Deisenroth2014a}, the authors consider the same dynamical system but solve different tasks by augmenting the policy with a task variable. In our work, we look at different settings of the dynamics but the task remains the same. We show how to generalize to the setting where task variables are latent and inferred from interaction data. This dramatically extends applicability in real-world settings.

\section{CONCLUSION}
We proposed a meta learning approach within the context of model-based RL that allows us to transfer knowledge from training configurations of robotic systems to unseen test configurations. The key idea behind our approach is to address the meta learning problem probabilistically using a latent variable model. We use online variational inference to obtain a posterior distribution over the latent variable, which describes the relatedness of tasks. This posterior is then used for long-term predictions of the state evolution and controller learning within a model-based RL setting. We demonstrated that our $\modelname$ approach is as efficient or better than a non-meta learning baseline when solving multiple tasks at once. The $\modelname$ further generalizes well to learning models and controllers for unseen tasks giving rise to substantial improvements in data-efficiency on novel tasks.

\subsubsection*{Acknowledgements}
This work was supported by Microsoft Research through its PhD Scholarship Programme.

\bibliography{refs}
\bibliographystyle{abbrv}

\end{document}